\documentclass{article} 

\usepackage{amsmath,amsfonts,bm}

\def\eqref#1{equation~\ref{#1}}

\def\1{\bm{1}}

\DeclareMathAlphabet{\mathsfit}{\encodingdefault}{\sfdefault}{m}{sl}
\SetMathAlphabet{\mathsfit}{bold}{\encodingdefault}{\sfdefault}{bx}{n}

\usepackage{iclr2026_conference,times}

\usepackage[utf8]{inputenc} 
\usepackage[T1]{fontenc}    
\usepackage{hyperref}       
\usepackage{url}            
\usepackage{booktabs}       
\usepackage[table]{xcolor}
\usepackage{amsfonts}       
\usepackage{nicefrac}       
\usepackage{microtype}      
\usepackage{xcolor}         

\usepackage{times}  
\usepackage{helvet}  
\usepackage{courier}  
\usepackage{graphicx} 
\usepackage{caption} 
\usepackage{amsfonts}
\usepackage{amsmath}
\usepackage[ruled,vlined]{algorithm2e}
\usepackage{fvextra}
\usepackage{newfloat}

\usepackage{xcolor}
\usepackage{listings}
\usepackage[most]{tcolorbox}
\tcbuselibrary{skins,breakable,listings}
\usepackage{caption}
\usepackage{fancyvrb}
\usepackage{amssymb}
\usepackage{wrapfig}

\usepackage{multirow}
\usepackage{makecell}
\usepackage{booktabs}
\usepackage{xspace}

\usepackage{booktabs}
\usepackage{multirow}
\usepackage{array}
\usepackage{colortbl}
\usepackage{pgf}

\newcolumntype{C}{>{\centering\arraybackslash}m{1cm}}

\newcommand{\heatcell}[1]{%
  \begingroup
  \pgfmathparse{min(max((#1-30)/(100-30),0),1)}%
  \edef\ratio{\pgfmathresult}%
  \pgfmathtruncatemacro{\shade}{round(8 + 82*\ratio)}%
  \edef\temp{\noexpand\cellcolor{barpurplebase!\shade}}%
  \temp #1%
  \endgroup
}

\usepackage{pifont}
\usepackage{enumitem}

\newcommand{\sys}{\texttt{MaR}\xspace}

\newcommand{\oss}{GPT-OSS-120B}
\newcommand{\xxlqwen}{Qwen3.5-397B}
\newcommand{\xlqwen}{Qwen3.5-122B}
\newcommand{\lqwen}{Qwen3.5-35B}
\newcommand{\ds}{Deepseek-V3.2}
\newcommand{\glm}{GLM-5.1}
\newcommand{\kimi}{Kimi-K2.5}

\newcommand{\mqwen}{Qwen3.5-9B}
\newcommand{\sqwen}{Qwen3.5-4B}

\definecolor{barpurplebase}{HTML}{C7B9E8}
\definecolor{softpurple}{HTML}{C7B3E5}
\definecolor{softgray}{HTML}{F5F5F7}

\definecolor{questionblue}{HTML}{2B8AC6}

\newcounter{finding}
\newcommand{\finding}[1]{%
    \refstepcounter{finding}%
    \begin{tcolorbox}[
        enhanced,
        breakable,
        colback=questionblue!6,
        colframe=questionblue!95,
        boxrule=0.75pt,
        arc=4pt,
        boxsep=3pt,
        left=4pt,
        right=4pt,
        top=4pt,
        bottom=4pt,
        before skip=8pt,
        after skip=8pt,
        drop shadow=gray!25!white
    ]
    \begin{center}
    #1
    \end{center}
    \end{tcolorbox}
}

\title{Metacognition as Reward:\\ Reinforcing LLM Reasoning via Knowledge and Regulation Signals}

\author{Sirui Chen\textsuperscript{1,2,3}\quad
        Lei Xu\textsuperscript{2,5}\quad
        Yuying Zhao\textsuperscript{6}\quad
        Yutian Chen\textsuperscript{1}\quad \\
        \textbf{Yu Wang\textsuperscript{1}\quad
        Beier Zhu\textsuperscript{4${\dag}$}\quad
        Hanwang Zhang\textsuperscript{3} \quad
        Shengjie Zhao\textsuperscript{1${\dag}$}\quad
        Chaochao Lu\textsuperscript{2${\dag}$}}
        \\
  $^{1}$Tongji University \quad
  $^{2}$Shanghai AI Laboratory \quad
  $^{3}$Nanyang Technological University \\
  $^{4}$University of Science and Technology of China \quad
  $^{5}$EPFL \quad
  $^{6}$Wuhan University \\
  \small \texttt{2111292@tongji.edu.cn}, \texttt{beier.zhu@ustc.edu.cn}, \texttt{luchaochao@pjlab.org.cn}\\
  }

\iclrfinalcopy

\begin{document}

\maketitle

\begin{NoHyper}
\def\thefootnote{\dag}\footnotetext{Corresponding author}
\end{NoHyper}

\begin{abstract}

Recent RL methods have substantially improved the reasoning abilities of LLMs. 
Existing reward designs mainly follow two paradigms: 
(1) Reinforcement learning with verifiable rewards (RLVR) derives outcome signals from executable checks or ground-truth answers, but provides limited guidance for intermediate reasoning behaviors. 
(2) Rubrics-as-reward (RaR) goes beyond final-answer checking by using natural-language rubrics to assess reasoning quality and task compliance, but often requires instance-specific rubrics and substantial design effort. 
To address these issues, we introduce \textbf{Metacognition-as-Reward} (\sys), a metacognition-inspired RL framework that guides LLM reasoning through two general process dimensions: 
i) metacognitive knowledge, which identifies task-relevant information without hand-crafted instance-specific rubrics, and 
ii) metacognitive regulation, which plans and adjusts the reasoning process to provide reward guidance beyond final-answer outcomes. 
\sys scaffolds model rollouts into explicit metacognitive components and optimizes them with a trajectory-level reward over task knowledge coverage, regulation fidelity, and final-answer correctness. 
In this way, \sys extends reward feedback to reasoning trajectories while grounding the reward signals in general metacognitive dimensions.
Experiments on 22 benchmarks show that \sys consistently improves model performance, achieving up to a 7.7\% gain over the base model and up to an 11.0\% gain over vanilla DAPO. Notably, \mqwen+\sys~narrows the gap to frontier models, surpassing \oss~on overall average and outperforming stronger models on several individual benchmarks.
Process-level analysis further shows substantial improvements in reasoning process quality. 
\sys also generalizes to out-of-domain datasets, where \sys-trained models improve over their corresponding base models on average.
Our code is available at \url{https://github.com/OpenCausaLab/MaR}.
\end{abstract}
\section{Introduction}

Large language models (LLMs) have shown strong capabilities in complex reasoning tasks, including mathematics, science, and coding~\citep{luo2025wizardmath,dai2026harder,shojaee2025llmsr,zhuo2025bigcodebench}. 
Recent reinforcement learning with verifiable rewards (RLVR) further improves LLM reasoning by optimizing models with outcome signals derived from verifiable answers or executable checks~\citep{shao2024deepseekmath,guo2025deepseek,lambert2025tulu,yu2025dapo,zhao2025unsupervised,zhao2026thinking}. 
However, such rewards leave the intermediate reasoning behaviors leading to the final answer weakly constrained~\citep{lightman2024lets,yu2026outcomerewardsguaranteeverifiable}.
Rubrics-as-reward (RaR) extends this paradigm by using natural-language rubrics to evaluate broader response qualities~\citep{zhou2025breaking,gunjal2026rubrics,wei2026qurl,li2026rubrichub}. 
Yet these rubrics are often tied to instance-specific criteria, requiring substantial design effort and sometimes human involvement, while providing limited guidance on what general reasoning dimensions should be encouraged across tasks. 
This leaves a central question:
\finding{
    \textit{Can we define more general reward objectives for guiding LLM reasoning?}
}

Metacognition provides a natural basis for defining such objectives. 
In cognitive psychology, metacognition is commonly understood as ``cognition about cognition'', encompassing knowledge of one's own cognitive processes and the ability to monitor and regulate them~\citep{flavell1979metacognition}.
It is commonly divided into \emph{metacognitive knowledge} and \emph{metacognitive regulation}~\citep{schraw1994assessing}. 
Metacognitive knowledge concerns awareness of task-relevant, available, and missing information, while metacognitive regulation concerns planning, monitoring, and adjusting the problem-solving process~\citep{schraw1998promoting}. 
Rather than prescribing task-specific criteria, this distinction characterizes higher-level cognitive processes in human problem solving and provides general reasoning dimensions that can be evaluated across tasks.

Based on this view, we propose \textbf{Metacognition-as-Reward} (\sys), a metacognition-inspired RL framework that consists of three components. 
(1) \emph{Metacognition-inspired rollout generation}: we scaffold model rollouts into an explicit structure that includes metacognitive knowledge, metacognitive regulation, optional lookback, and the final answer. 
Metacognitive knowledge guides the model to identify task-relevant information, while metacognitive regulation guides it to plan and track the reasoning process. 
When uncertainty arises, the model can use lookback to recover missing information before producing the answer. 
(2) \emph{Metacognitive monitoring reward}: we define a trajectory-level reward over this structure that evaluates knowledge coverage, regulation fidelity, and final-answer correctness. 
This reward goes beyond task success by evaluating task-specific reasoning content through general metacognitive dimensions. 
(3) \emph{Policy optimization}: we optimize the policy model using this reward, encouraging responses that are both task-correct and process-aligned.

We conduct comprehensive experiments on 22 benchmarks spanning science, medicine, long-context reasoning, mathematical reasoning, and logical reasoning. 
\sys achieves up to a 7.7\% improvement over the base model on the science benchmark, and \mqwen+\sys surpasses \oss~and \ds~(685B) on selected medical benchmark.
Compared with vanilla DAPO, \sys obtains up to an 11.0\% gain on the rubric-based benchmark, suggesting that metacognitive supervision improves reasoning quality beyond final-answer optimization. 
Process-level analysis on long-context tasks further shows average improvements of 17.2\%, 10.7\%, and 9.8\% in knowledge monitoring, regulation fidelity, and final-answer correctness, respectively. 
\sys also generalizes to out-of-domain (OOD) mathematical and logical reasoning benchmarks, where the 9B model achieves an average score of 75.1\%, outperforming the base model by 2.4\%.

To summarize, our main contributions are as follows:
\begin{itemize}[leftmargin=*]
\item We formulate metacognition as a general principle for reward design, highlighting knowledge monitoring and reasoning regulation as reward dimensions beyond final-answer correctness.

\item We propose \sys, a metacognition-inspired RL framework that scaffolds model rollouts into metacognitive knowledge, metacognitive regulation, and final answer, and optimizes them with a trajectory-level reward over knowledge coverage, regulation fidelity, and answer correctness.

\item We validate \sys on 22 benchmarks across five domains. Extensive results show consistent performance gains, improved metacognitive process scores, and robust OOD generalization.
\end{itemize}

\section{Related Work}

\noindent\textbf{Metacognitive reasoning in LLMs.}
Recent work studying metacognition in LLMs can be broadly categorized into three perspectives: understanding whether such abilities exist, eliciting them through prompting, and improving them through training.
For analysis, \citet{ma-etal-2025-large-language-models} study how to measure LLM meta-cognition more reliably, while \citet{ji-an2026language} examine whether LLMs exhibit metacognitive monitoring and control over their internal activations.
These studies suggest that LLMs show partial metacognitive abilities, but such abilities remain unstable.
At inference time, \citet{wang-zhao-2024-metacognitive} introduce a structured self-evaluation procedure inspired by human introspective reasoning.
\citet{lee-etal-2025-pragmatic} combine pragmatic reasoning with reflective prompting for sarcasm detection, and \citet{ta2025mdtoc} apply metacognitive prompting to mathematical reasoning through concept-level decomposition, calculation verification, and solution comparison.
However, these methods do not internalize metacognitive abilities into the model.
For RL, \citet{wan2026rema} use multi-agent RL to separate high-level meta-thinking from low-level reasoning execution.
\citet{ha2025aha} decouple reasoning from control to regulate emergent behaviors such as reflection and backtracking.
\citet{sun2025cog} improve sample efficiency with a hierarchical metacognitive RL framework based on decomposition and refinement.
These studies show that RL can induce metacognitive behaviors.
However, they mainly study how to elicit or organize such behaviors, leaving underexplored what constitutes a high-quality metacognitive process during reasoning and how to learn it as a general training objective.

\noindent\textbf{Reward design for LLM reasoning.}
Recent RL methods for LLM reasoning largely rely on outcome rewards derived from verifiable answers or executable checks.
Such rewards are effective in tasks where correctness can be automatically judged~\citep{team2025kimi,zeng2025simplerlzoo,hu2026openreasonerzero,wang2026reinforcement,zhao2026absolute,wang2026textbfre}.
Despite their effectiveness, verifiable outcome rewards provide limited guidance on intermediate reasoning behaviors and are less applicable to open-generation tasks, where correctness cannot always be reduced to a single verifiable answer.
RaR extend reward design beyond verifiable outcomes by using natural-language criteria to evaluate broader response qualities.
They can capture dimensions such as helpfulness, completeness, reasoning quality, and task-specific constraints~\citep{liu2025openrubrics,xie2025auto,shen2026rethinking,ye2025self,masters2026arcane}.
However, these objectives are often derived from instance-specific criteria, and thus provide limited guidance on what general reward objectives should guide LLM reasoning across tasks.

\section{Method}
\begin{figure}[t!]  
\centering
\includegraphics[width=.95\columnwidth]{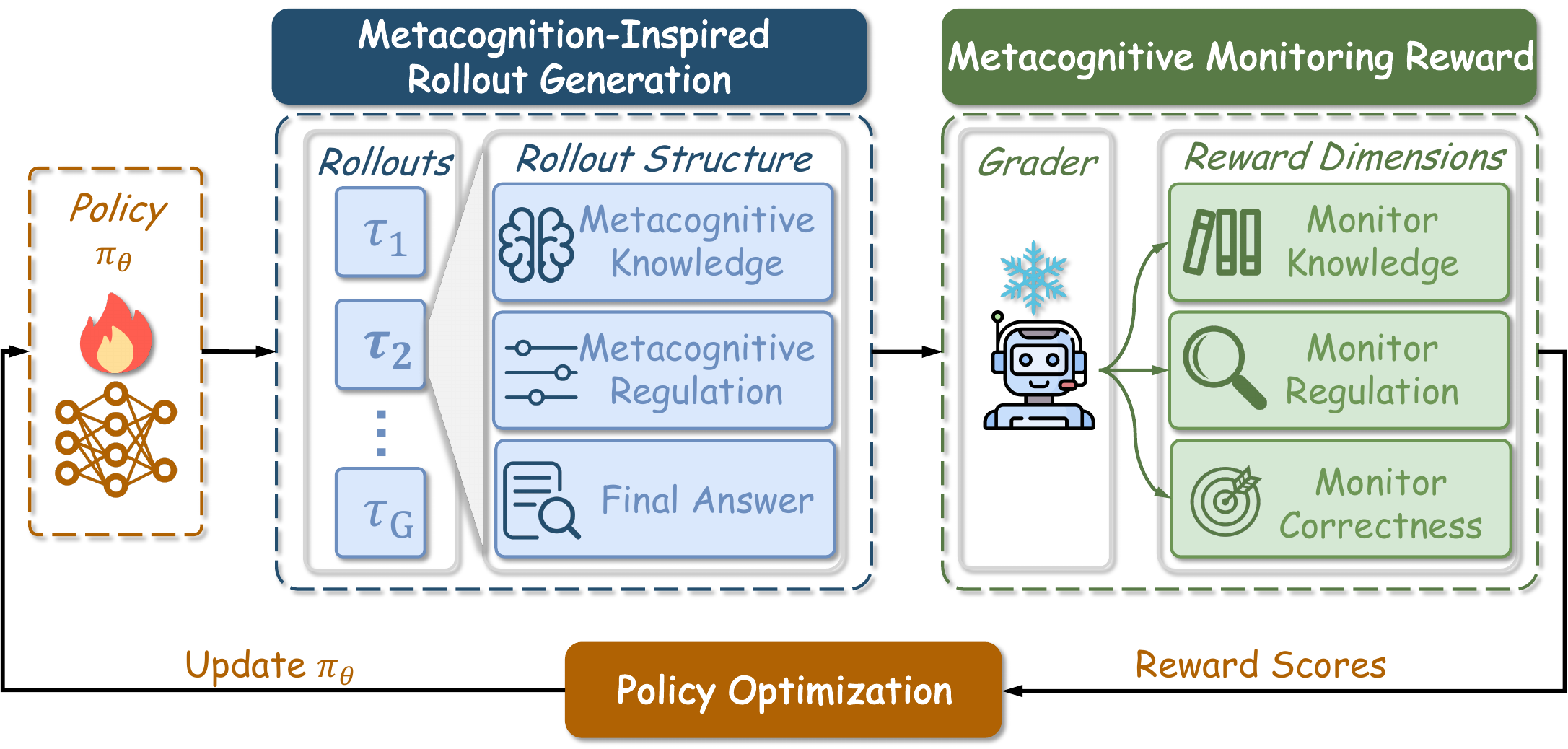}
\caption{
\textbf{Overview of \sys.}
\sys follows a three-stage loop: the policy generates multiple structured rollouts consisting of metacognitive knowledge, metacognitive regulation, and the final answer; a grader scores each rollout along knowledge monitoring, regulation monitoring, and answer correctness; and the resulting reward scores are used to optimize the policy.
}
\label{fig:main_method}
\end{figure}

We instantiate this metacognitive view as a structured RL framework, as shown in Figure~\ref{fig:main_method}. 
The core idea is to make intermediate reasoning behaviors observable and rewardable during policy optimization. 
Specifically, each response is organized into metacognitive knowledge (MK), metacognitive regulation (MR), optional \texttt{LOOKBACK}, and a final answer. 
This structure reveals which task-relevant information the model identifies, how it plans and monitors its reasoning, and whether it recovers missing information under uncertainty.
We next formalize the learning objective (Section~\ref{sec:formulation}), introduce the structured rollout format (Section~\ref{sec:generation}), define the metacognitive reward (Section~\ref{sec:reward}), and describe policy optimization (Section~\ref{sec:optimization}).

\subsection{Problem Formulation}
\label{sec:formulation}
Formally, let $\mathcal{D}$ be a training set of input-answer pairs $(x,y)$, where $x$ denotes the input instance and $y$ denotes the ground truth. 
Each instance is also annotated with a fixed set of gold knowledge units, which specifies the task-relevant knowledge required for solving the problem.
Our goal is to optimize a policy $\pi_\theta$ to generate responses that are both process-aligned and task-correct. 
For each $x$, $\pi_\theta$ generates a structured rollout $\tau \sim \pi_\theta(\cdot \mid x)$.
We define a metacognition-inspired reward $\mathsf{R}(\tau,x,y)$ that evaluates the quality of MK, the consistency of MR-guided execution, and the correctness of the final answer. 
The high-level objective is:
\begin{equation}
\max_{\theta}\;
\mathbb{E}_{(x,y)\sim\mathcal{D},\;\tau\sim\pi_\theta(\cdot\mid x)}
\left[
\mathsf{R}(\tau,x,y)
\right].
\end{equation}

\subsection{Metacognition-Inspired Rollout Generation}
\label{sec:generation}
\begin{wrapfigure}{r}{0.5\columnwidth}
\vspace{-13pt}
\centering
\includegraphics[width=\linewidth]{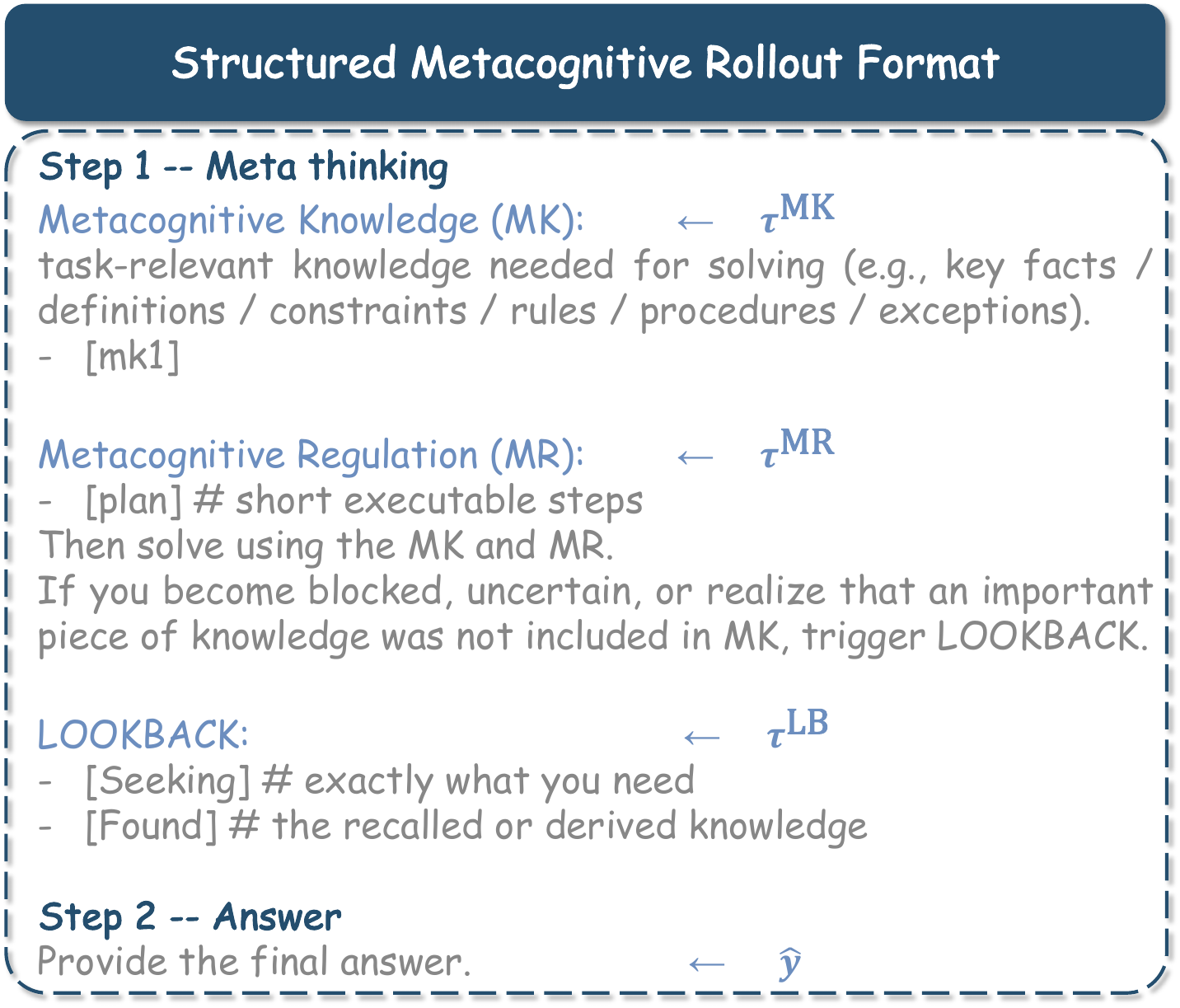}
\caption{
\textbf{Key components of the rollout generation prompt.}
The prompt instructs the policy to organize each rollout into MK, MR, an optional \texttt{LOOKBACK}, and the final answer.
}
\vspace{-3pt}
\label{fig:prompt_demo}
\end{wrapfigure}

We operationalize the metacognitive structure by prompting the current policy $\pi_\theta$ to generate each rollout in four parts:
\begin{equation}
\tau = (\tau^{\mathrm{MK}}, \tau^{\mathrm{MR}}, \tau^{\mathrm{LB}}, \hat{y}),
\end{equation}
where $\tau^{\mathrm{MK}}$ denotes metacognitive knowledge, $\tau^{\mathrm{MR}}$ denotes metacognitive regulation, $\tau^{\mathrm{LB}}$ denotes an optional \texttt{LOOKBACK}, and $\hat{y}$ is the final answer. 
Figure~\ref{fig:prompt_demo} shows the key components of the prompt used to elicit structured rollouts from the policy $\pi_\theta$, with the full prompt provided in Figure~\ref{fig:appendix_rollout}. 
Figure~\ref{fig:appendix_policy} presents an example rollout generated by $\pi_\theta$.

\noindent \textbf{Metacognitive knowledge.}
The MK part asks the policy to explicitly identify the task-relevant knowledge needed for solving the problem, including key facts, definitions, constraints, rules, procedures, exceptions, etc.
To make knowledge coverage easier to inspect and evaluate, we require MK to be written as atomic indexed items, e.g., \texttt{[mk1]}, \texttt{[mk2]}, and so on.
This design makes knowledge monitoring explicit before the model proceeds to reasoning and answer generation.

\noindent \textbf{Metacognitive regulation.}
After identifying the relevant knowledge, the policy specifies a \texttt{[plan]} that describes how it will solve the problem.
The plan is written as short, executable steps, so that the subsequent reasoning has an explicit procedure to follow.
This structure exposes the model's regulation process and supports subsequent evaluation of regulation fidelity.

\noindent \textbf{\texttt{LOOKBACK}.}
This is a recovery step triggered when the policy becomes blocked, uncertain, or detects that some necessary knowledge was omitted from MK.
It first states the missing information in \texttt{[Seeking]}, and then records the recalled or derived knowledge in \texttt{[Found]}.
This step makes knowledge recovery explicit during reasoning, and allows the grader to assess whether the policy can correct an incomplete initial knowledge state.

\subsection{Metacognitive Monitoring Reward}
\label{sec:reward}
Inspired by the monitoring-control framework \citep{nelson1990metamemory}, we design a metacognitive monitoring reward ($\mathsf{MMR}$) to evaluate whether $\pi_\theta$ can: i) accurately identify task-relevant knowledge, ii) appropriately regulate its subsequent reasoning, and iii) produce a correct final answer.
For each rollout, $\mathsf{MMR}$ is computed from fixed dataset annotations and grader-estimated rollout signals.
The fixed annotations provide the gold knowledge units for each instance, while an LLM grader $\pi_{\mathsf{G}}$ estimates the rollout-dependent quantities required by the component rewards defined below.
The full grader prompt and scoring protocol are provided in Figure~\ref{fig:appendix_grader}.

\noindent\textbf{Knowledge monitoring reward ($\mathsf{KMR}$).}
We evaluate whether the policy $\pi_\theta$ identifies the task-relevant knowledge required for solving the problem from two aspects: (1) the initial coverage of gold knowledge units in MK, and (2) the recovery of initially missed knowledge through \texttt{LOOKBACK}. 

Let $n$ denote the total number of gold knowledge units annotated for the instance.
Given the policy rollout, the grader $\pi_{\mathsf{G}}$ estimates $k$, the number of gold knowledge units correctly identified in the initial MK.
It also estimates $r$, the number of gold knowledge units that are missed in the initial MK but successfully recovered through \texttt{LOOKBACK}.
If \texttt{LOOKBACK} is not used, we set $r=0$.
We define the knowledge monitoring reward as:
\begin{equation}
\mathsf{KMR}=\tfrac{k+r}{n}.
\end{equation}
Thus, $\mathsf{KMR}$ measures the proportion of required knowledge units covered either by the initial MK or by subsequent \texttt{LOOKBACK}.

\noindent\textbf{Regulation monitoring reward ($\mathsf{RMR}$).}
Beyond knowledge identification, we also monitor whether $\pi_\theta$ follows its self-generated MR during subsequent reasoning.
Concretely, we define the regulation monitoring reward as:
\begin{equation}
\mathsf{RMR}=a(1-\lambda s),
\label{eq:rmr}
\end{equation}
where $a\in[0,1]$ is estimated by the grader $\pi_{\mathsf{G}}$ and measures the consistency between the policy's actual reasoning process and its stated MR, $s\in\{0,1\}$ is a grader-estimated shortcut indicator, and $\lambda\in[0,1]$ is a shortcut penalty coefficient.
A shortcut occurs when the policy reaches the final answer while bypassing its own declared regulatory process.
Thus, $a$ rewards faithful plan execution, while the multiplicative penalty discourages such shortcut trajectories.
Since intermediate reasoning traces are not always faithfully translated into final answers \citep{lanham2023measuring,paul-etal-2024-making}, this process-level consistency serves as a useful supervision signal beyond answer correctness alone.

\noindent\textbf{Correctness reward ($\mathsf{CR}$).}
We define a task-level reward to evaluate whether the final answer $\hat{y}$ matches the ground truth $y$:
\begin{equation}
\mathsf{CR}(\hat{y}, y)
=
\mathbb{I}\bigl(\hat{y}=y \ \lor\  \hat{y}\simeq y\bigr),
\end{equation}
where $\mathbb{I}(\cdot)$ is the indicator function, and $\simeq$ denotes semantic equivalence judged by the grader $\pi_{\mathsf{G}}$.
Thus, $\mathsf{CR}=1$ if the answer either exactly matches the ground truth or is judged semantically equivalent to it, and $\mathsf{CR}=0$ otherwise.

\noindent\textbf{Final reward.}
We define the final reward as the sum of the three components:
\begin{equation}
\mathsf{R}=\mathsf{KMR}+\mathsf{RMR}+\mathsf{CR}.
\end{equation}
Here, $\mathsf{KMR}$, $\mathsf{RMR}$, and $\mathsf{CR}$ respectively capture $\pi_\theta$'s MK, MR, and final answer correctness, with each term bounded in $[0,1]$.


\subsection{Policy Optimization}
\label{sec:optimization}

We optimize $\pi_\theta$ with DAPO \citep{yu2025dapo}. For each input $x$, we sample a group of $G$ rollouts $\{\tau_i\}_{i=1}^{G}$ from the current policy and compute their rewards $\{\mathsf{R}_i\}_{i=1}^{G}$ using the metacognitive reward defined above. Following group-relative policy optimization, we estimate the advantage of each rollout by normalizing rewards within the same group:
\begin{equation}
\hat{A}_i =
\frac{
\mathsf{R}_i-\mathrm{mean}(\{\mathsf{R}_j\}_{j=1}^{G})
}{
\mathrm{std}(\{\mathsf{R}_j\}_{j=1}^{G})+\delta
},
\end{equation}
where $\delta$ is a constant for numerical stability.
We then update the policy using the DAPO objective:
\begin{equation}
\mathcal{J}_{\mathrm{DAPO}}(\theta)
=
\mathbb{E}
\left[
\frac{1}{\sum_{i=1}^{G}|\tau_i|}
\sum_{i=1}^{G}
\sum_{t=1}^{|\tau_i|}
\min\left(
\rho_{i,t}(\theta)\hat{A}_i,\;
\mathrm{clip}\!\left(
\rho_{i,t}(\theta),
1-\epsilon_{\mathrm{low}},
1+\epsilon_{\mathrm{high}}
\right)\hat{A}_i
\right)
\right],
\end{equation}
where $\rho_{i,t}(\theta)$ denotes the probability ratio used in the clipped policy objective, $\epsilon_{\mathrm{low}}$ and $\epsilon_{\mathrm{high}}$ are the lower and upper clipping coefficients. This objective increases the likelihood of rollouts with higher metacognitive rewards while suppressing rollouts with incomplete knowledge identification, inconsistent regulation, or incorrect final answers.
\section{Experiment}
\label{sec:experiment}

\subsection{Setup}
\label{sec:setups}

\noindent\textbf{Baselines.}
We evaluate a diverse suite of frontier models across various developers and scales as baselines, including Qwen3.5-\{4B,9B,35B,122B,397B\} \citep{qwen3.5}, \oss~\citep{openai2025gptoss120bgptoss20bmodel}, \ds~(685B) \citep{liu2025deepseek}, \glm~(754B) \citep{zeng2026glm}, and \kimi~(1.1T) \citep{team2026kimi}. In terms of training paradigm, we compare our method with DAPO \citep{yu2025dapo}. All models are evaluated with zero-shot inference, where each model is directly given the question and asked to produce an answer.

\noindent\textbf{Training settings.}
We post-train both \sqwen~and \mqwen~using the DAPO algorithm implemented within the VeRL framework \citep{10.1145/3689031.3696075}. Our training data are derived from RaR-Medicine and RaR-Science \citep{gunjal2026rubrics}. We use GPT-5.1 to generate gold knowledge, resulting in approximately 32K training samples. The annotation prompt is provided in Figure \ref{fig:data_generation}. For each training sample, we draw 8 responses from the current policy $\pi_\theta$ and use \xxlqwen~as the judge model to provide rewards. The $\lambda$ in Equation \ref{eq:rmr} is set to 0.3. More details are provided in Section \ref{appendix_training_details}.

\noindent\textbf{Datasets.}
As shown in Figure~\ref{fig:dataset}, we conduct a thorough evaluation of \sys on 22 datasets: 

\begin{wrapfigure}{l}{0.6\columnwidth}
\vspace{-10pt}
\centering
\includegraphics[width=\linewidth]{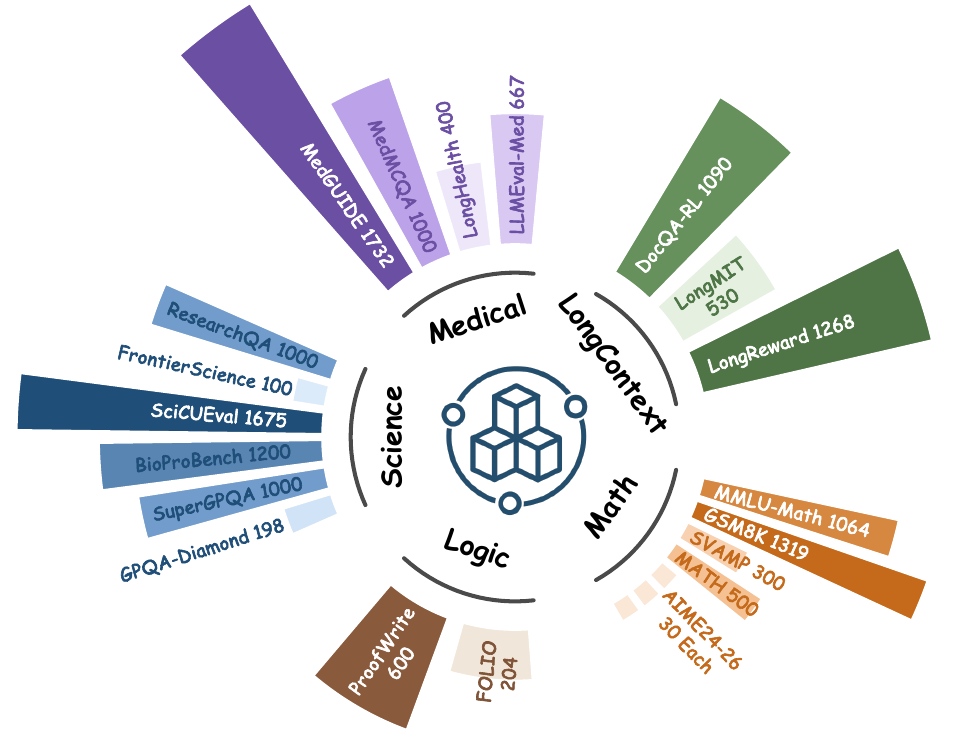}
\caption{
\textbf{Overview of the evaluation datasets.}
}
\vspace{-8pt}
\label{fig:dataset}
\end{wrapfigure}
Six in science, four in medical, three in long-context reasoning, seven in mathematical reasoning, and two in logical reasoning. These datasets contain approximately 16K test examples in total, providing broad task coverage and supporting reliable evaluation.
(1) Science datasets: 
GPQA-Diamond \citep{rein2024gpqa},
SuperGPQA \citep{du2025supergpqa},
BioProBench \citep{liu2025bioprobench},
SciCUEval \citep{yu2026scicueval},
FrontierScience \citep{wang2026frontierscience}, and
ResearchQA \citep{yifei2025researchqa}.
(2) Medical datasets:
MedGUIDE \citep{li2025medguide},
MedMCQA \citep{pmlr-v174-pal22a},
LongHealth \citep{adams2025longhealth}, and
LLMEval-Med \citep{zhang-etal-2025-llmeval}.
(3) Long-context reasoning datasets:
DocQA-RL-1.6K \citep{wan2025qwenlong},
LongMIT \citep{chen-etal-2025-essential}, and
LongReward \citep{zhang-etal-2025-longreward}.
(4) Mathematical reasoning datasets:
MMLU-Math \citep{hendrycks2021measuring},
GSM8K \citep{cobbe2021training},
SVAMP \citep{patel-etal-2021-nlp},
MATH-500 \citep{hendrycks2021measuring2},
AIME 2024, 2025 and 2026 \citep{balunovic2025matharena}.
(5) Logical reasoning datasets: 
FOLIO \citep{han-etal-2024-folio} and ProofWriter \citep{tafjord-etal-2021-proofwriter}.

\subsection{Main Results}
\label{sec:main_results}

\begin{table*}[t]
\centering
\setlength{\tabcolsep}{2.8pt}
\fontsize{8}{10}\selectfont

\caption{
\textbf{Broad evaluation across science and medical benchmarks.}
We compare \sys with base models, vanilla DAPO, and frontier models.
The table reports average accuracy on each dataset, covering both outcome-based and rubric-based evaluations. Best results in each group are in \textbf{bold}.
}
\begin{tabular}{lcccccccccc}
\specialrule{2pt}{0pt}{0pt}
\noalign{\vskip 2pt}

\multirow{2}{*}{\textbf{Model}} &
\multicolumn{6}{c}{\textbf{Science}} &
\multicolumn{4}{c}{\textbf{Medical}} \\
\cmidrule(lr){2-7}
\cmidrule(lr){8-11}

& GPQA-D & S-GPQA & 
BioPro & SciCU &
F-Science & R-QA & 
MedGUIDE & MedMCQA & 
L-Health &L-Med\\
\midrule

\rowcolor{gray!20} \multicolumn{11}{c}{\textbf{Frontier Models} \rule{0pt}{6pt}} \\

\lqwen   & 71.2 & 49.7 & 67.3 & 94.7 & 54.3 & 78.3 & 43.0 & 79.2 & 89.8 & 72.7 \\
\oss     & 65.7 & 46.9 & 63.5 & 94.9 & 34.4 & 85.8 & 44.1 & 74.9 & 87.5 & 62.2 \\
\xlqwen  & 73.7 & 53.0 & 69.2 & 94.9 & 59.8 & 79.0 & 48.7 & 81.8 & 90.2 & 74.5 \\
\xxlqwen & 73.3 & 55.4 & 69.4 & 96.2 & 70.5 & 79.4 & 49.9 & 83.2 & 90.8 & 75.7 \\
\ds      & 77.0 & 54.6 & 68.5 & 95.6 & 54.2 & 81.5 & 46.0 & 80.7 & 89.2 & 72.5 \\
\glm     & 67.5 & 48.9 & 67.5 & 95.9 & 38.2 & 80.5 & 45.8 & 81.3 & 89.7 & 73.9 \\
\kimi    & 76.4 & 57.3 & 70.0 & 97.1 & 55.2 & 84.9 & 47.8 & 82.3 & 90.4 & 74.6 \\

\midrule

\rowcolor{gray!20} \multicolumn{11}{c}{\textbf{Ours} \rule{0pt}{6pt}} \\

\sqwen & 63.3 & 38.0 & 58.0 & 93.0 & \textbf{34.7} & 72.2 & 28.9 & 67.5 & 86.8 & 62.4 \\

\quad + DAPO & 65.9 & 40.6 & 58.5 & 93.5 & 33.0 & 70.4 & \textbf{30.5} & 67.6 & 87.4 & 62.1 \\

\rowcolor{teal!15} \quad + \sys & \textbf{68.7} & \textbf{41.7} & \textbf{58.8} & \textbf{94.9} & 34.0 & \textbf{72.7} & 30.3 & \textbf{69.0} & \textbf{88.5} & \textbf{62.9} \\


\cmidrule(lr){1-11}

\mqwen & 67.0 & 43.3 & 61.5 & 93.8 & 42.4 & 74.6 & 36.6 & 73.4 & 87.1 & 66.7 \\

\quad + DAPO & 71.1 & 44.8 & 61.7 & 94.4 & 39.0 & 73.4 & 37.6 & 73.5 & 88.0 & 66.3 \\

\rowcolor{teal!15} \quad + \sys & \textbf{74.7} & \textbf{46.4} & \textbf{62.8} & \textbf{95.5} & \textbf{50.0} & \textbf{76.0} & \textbf{38.5} & \textbf{74.2} & \textbf{89.3} & \textbf{68.7} \\


\specialrule{2pt}{0pt}{0pt}
\end{tabular}
\label{tab:main_results}
\end{table*}

Table~\ref{tab:main_results} presents a detailed comparison between \sys~and various baselines. We find that:

\noindent\textbf{\sys~consistently improves broad performance across science and medical benchmarks.}
Across the ten benchmarks, \sys~improves the performance of \sqwen~on 9 out of 10 benchmarks, with the largest gain observed on GPQA-Diamond (+5.4\%).
The improvement becomes more pronounced on the stronger \mqwen~backbone, where \mqwen+\sys~outperforms \mqwen~on all ten benchmarks.
The largest gains over \mqwen~and DAPO reach 7.7\% and 11.0\%, respectively.
These results show that \sys~provides broad and stable gains across complex reasoning domains.
Moreover, \sys~outperforms vanilla DAPO, showing that the gains come from metacognitive supervision rather than generic RL alone.
For \mqwen, DAPO improves the overall average from 64.6\% to 65.0\%, an absolute improvement of only 0.4\%, whereas \sys~raises it to 67.6\%.

\noindent\textbf{Stronger base models benefits more from \sys and narrows the gap to frontier models.}
\sys~brings a larger gain on \mqwen~than on \sqwen.
The same trend holds within each domain: on science benchmarks, the improvement increases from 1.9\% for \sqwen~to 3.8\% for \mqwen; on medical benchmarks, it increases from 1.3\% to 1.7\%.
This suggests that stronger base models can better benefit from metacognitive supervision.
Moreover, \mqwen+\sys~narrows the gap to frontier models.
Its overall average reaches 67.6\%, surpassing \oss~and approaching \glm.
On individual benchmarks, \mqwen+\sys~outperforms \xxlqwen~on GPQA-Diamond, \xlqwen~on SciCUEval, and \ds~on LongHealth.

\noindent\textbf{\sys~improves the quality of reasoning processes via metacognitive supervision.}
The advantage of \sys~is especially evident on FrontierScience, ResearchQA, and LLMEval-Med, where evaluation relies on rubrics rather than only final-answer correctness.
Since DAPO uses final-answer correctness as the reward, it provides limited supervision for intermediate reasoning quality.
This is reflected in its performance drops on these benchmarks: for \mqwen, DAPO decreases the scores by 3.4\%, 1.2\%, and 0.4\%, respectively.
In contrast, \sys~provides explicit supervision for metacognitive knowledge and regulation, leading to better process-oriented reasoning quality.
Compared with DAPO, \sys~improves \mqwen~by 11.0\%, 2.6\%, and 2.4\%, respectively.
The average score on these benchmarks drops from 61.2\% to 59.6\% under DAPO, but increases to 64.9\% with \sys.

\subsection{Mechanism Analysis}
\begin{figure}[t!]  
\centering
\includegraphics[width=\columnwidth]{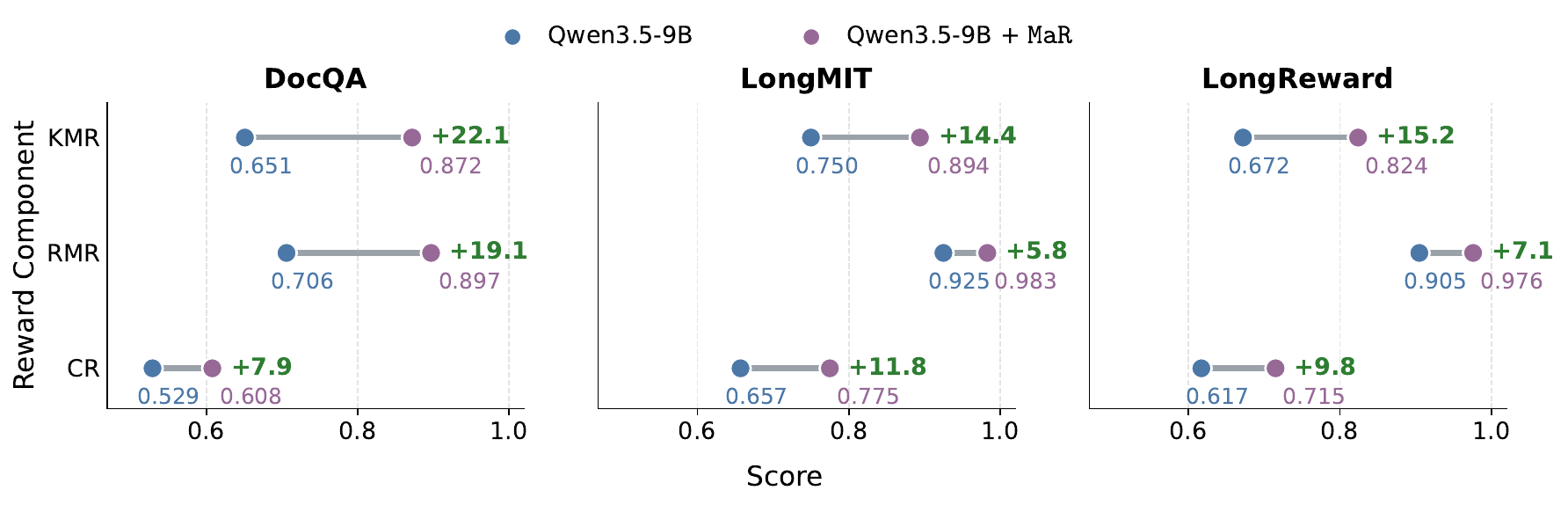}
\caption{
\textbf{Metacognitive process scores on long-context reasoning tasks.}
An external grader scores the generated metacognitive process along three reward components: KMR, RMR, and CR.
The \textcolor[HTML]{2E7D32}{green annotations} indicate the absolute improvements (\%) brought by \sys.
}
\label{fig:component_dumbbell}
\end{figure}

Figure~\ref{fig:component_dumbbell} evaluates whether \sys improves the model's metacognitive process. 
Specifically, we evaluate \mqwen~and \mqwen+\sys~on DocQA, LongMIT, and LongReward, where no long-context reasoning data is used during training. 
At inference time, both models use the prompt in Figure \ref{fig:appendix_rollout}, and we use \xxlqwen~to score the generated metacognitive process along three reward components: KMR, RMR, and CR. 
Our key findings are as follows:

\noindent\textbf{\sys~generalizes metacognitive process improvements to OOD long-context reasoning tasks.}
Across all three datasets, \sys~consistently improves all reward components over the base model.
On DocQA, KMR, RMR, and CR increase by 22.1\%, 19.1\%, and 7.9\%, respectively.
On LongMIT, the corresponding improvements are 14.4\%, 5.8\%, and 11.8\%, while on LongReward they are 15.2\%, 7.1\%, and 9.8\%.
These consistent gains suggest that \sys~does not merely fit the training distribution, but learns transferable metacognitive behaviors for unseen long-context reasoning scenarios.

\noindent\textbf{The improvement is primarily reflected in process-level metacognition and is accompanied by better correctness.}
Averaged across the three OOD datasets, \sys~improves KMR by 17.2\% and RMR by 10.7\%, indicating that the model becomes better at identifying relevant knowledge and regulating its reasoning process.
CR also improves by 9.8\% on average, showing that better metacognitive processes are accompanied by improved final correctness.
This provides evidence that \sys~improves reasoning quality beyond merely optimizing final-answer accuracy.

Figure~\ref{fig:component_heatmap} further analyzes the relationship among KMR, RMR, and CR by computing their pairwise Spearman correlations \citep{spearman1904proof}. We find that:

\begin{wrapfigure}{r}{0.7\columnwidth}
\vspace{-8pt}
\centering
\includegraphics[width=\linewidth]{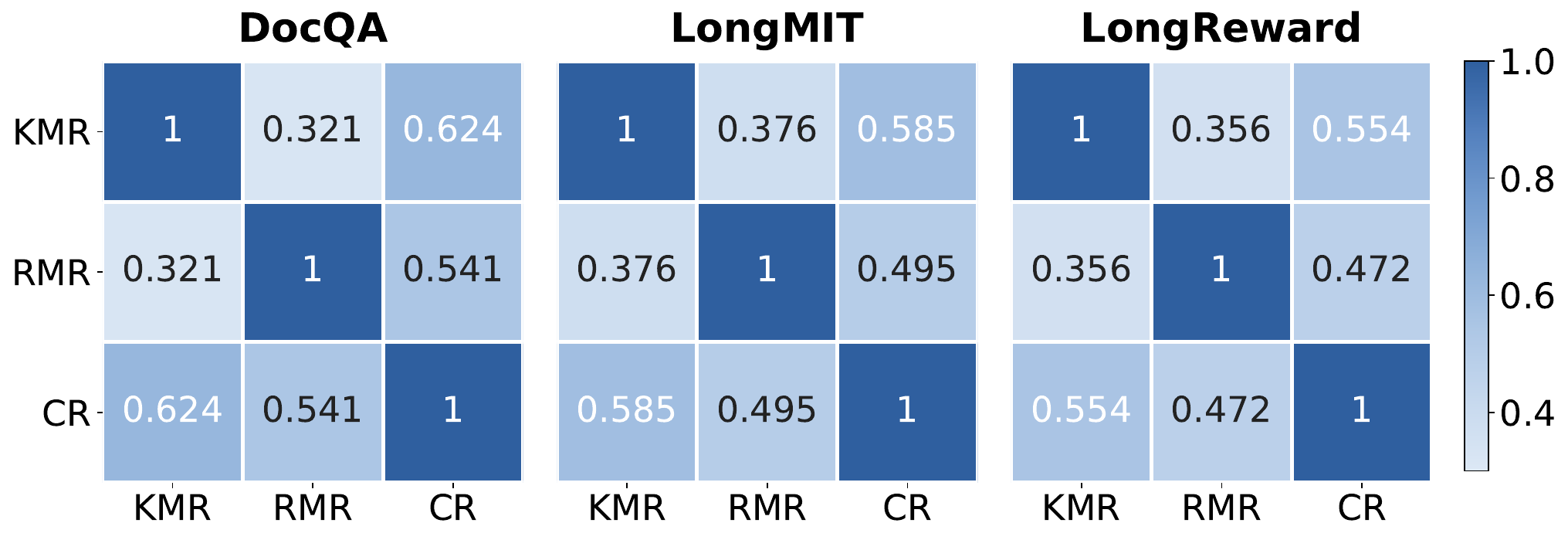}
\caption{
\textbf{Spearman correlations among metacognitive reward components.}
KMR and RMR capture complementary aspects of the reasoning process, while CR measures final-answer correctness.
}
\vspace{-8pt}
\label{fig:component_heatmap}
\end{wrapfigure}

\noindent\textbf{KMR and RMR are both positively associated with final correctness.}
Across three datasets, KMR shows consistently high correlations with CR, with Spearman coefficients of 0.624, 0.585, and 0.554, respectively.
RMR is also positively correlated with CR, with coefficients of 0.541, 0.495, and 0.472.
This suggests that both knowledge monitoring and regulation fidelity are closely related to final correctness, supporting the motivation of supervising these two metacognitive components.

\noindent\textbf{KMR and RMR provide non-redundant supervision.}
The correlations between KMR and RMR are moderate, with coefficients of 0.321, 0.376, and 0.356 across the three datasets.
This indicates that they are related, since identifying task-relevant knowledge can support better planning and execution.
However, these correlations are lower than their respective correlations with CR, suggesting that they capture distinct aspects of the reasoning process rather than duplicating each other.

\subsection{Generalizability}
\begin{figure}[t!]  
\centering
\includegraphics[width=.9\columnwidth]{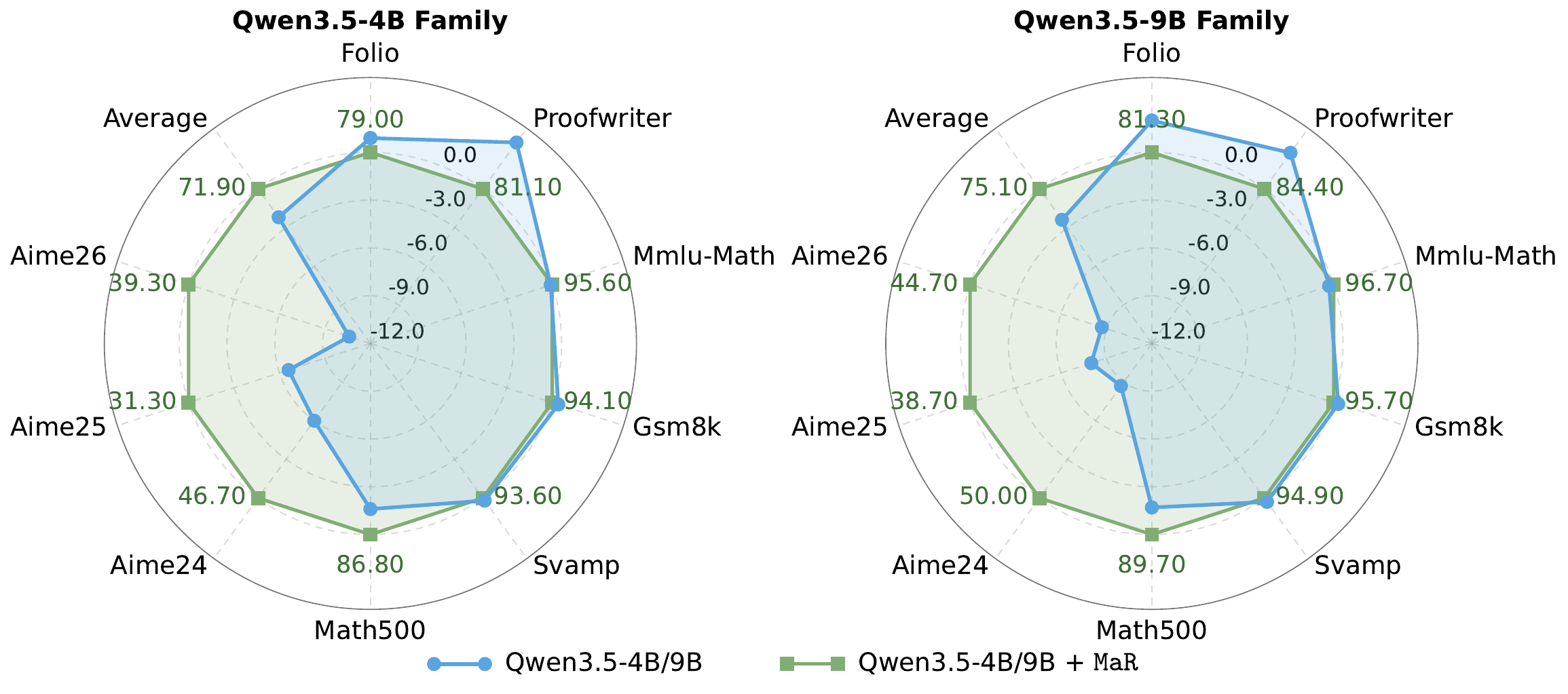}
\caption{
\textbf{Model generalizability across math and logical reasoning benchmarks.}
We take the Qwen3.5-4B/9B+\sys as the reference baseline in each panel.
The \textcolor[HTML]{80AD74}{green squares} mark the Qwen3.5-4B/9B+\sys's absolute performance.
The \textcolor[HTML]{5AA4E0}{blue circles} show the corresponding baseline model, positioned according to the performance difference.
Negative radial values indicate that the baseline model performs worse than its \sys-trained counterpart.
}
\label{fig:ood_radar}
\end{figure}

Figure~\ref{fig:ood_radar} evaluates \sys on seven math reasoning datasets and two logical reasoning OOD datasets. 
We conclude that:


\noindent\textbf{\sys demonstrates strong generalizability beyond the training domains.}
\sys improves the average performance of both \sqwen~and \mqwen~on OOD reasoning benchmarks.
For \sqwen, the average score increases from 69.7\% to 71.9\%, while for \mqwen, it increases from 72.7\% to 75.1\%.
The gains are especially clear on more challenging mathematical reasoning tasks.
For \sqwen, \sys improves AIME-2024, AIME-2025, and AIME-2026 by 6.0\%, 6.6\%, and 10.6\%, respectively.
For \mqwen, the corresponding gains are 8.7\%, 8.0\%, and 8.7\%.
\sys also improves MATH500 by 1.6\% on \sqwen~and 1.7\% on \mqwen.
These results suggest that metacognitive supervision transfers most effectively to tasks requiring complex reasoning and self-regulation.
Although performance slightly decreases on FOLIO and ProofWriter, the overall gains indicate that \sys learns transferable reasoning behaviors rather than domain-specific answers.

\subsection{Ablation Studies}
\label{sec:ablation_studies}


\begin{figure*}[t!]
\centering
\begin{minipage}[t]{0.49\textwidth}
\centering
\includegraphics[width=\linewidth]{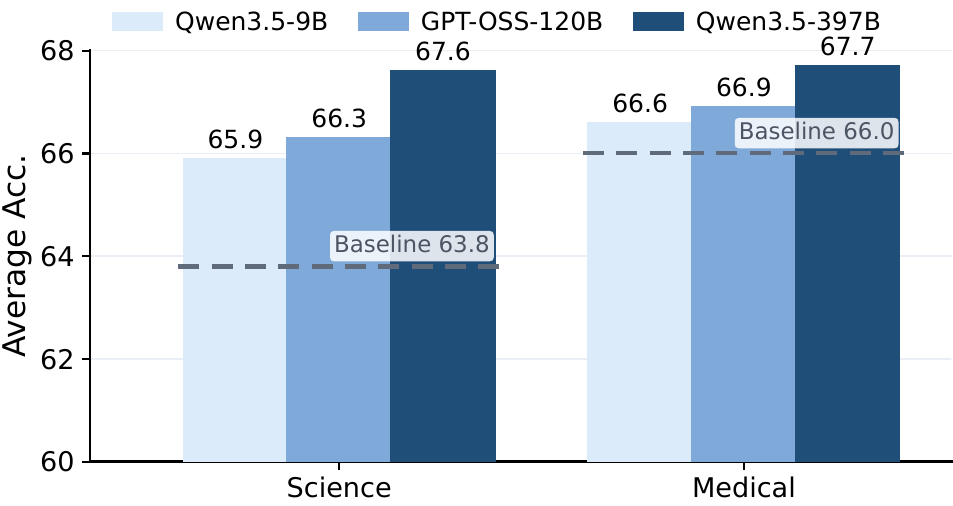}
\caption{
\textbf{Effect of grader choice.}
We train the \mqwen~with \sys using varying sizes of graders.
The figure reports average accuracy on science and medical benchmarks, with dashed lines indicating the \mqwen~baseline.
}
\label{fig:ablation_grader}
\end{minipage}
\hfill
\begin{minipage}[t]{0.49\textwidth}
\centering
\includegraphics[width=\linewidth]{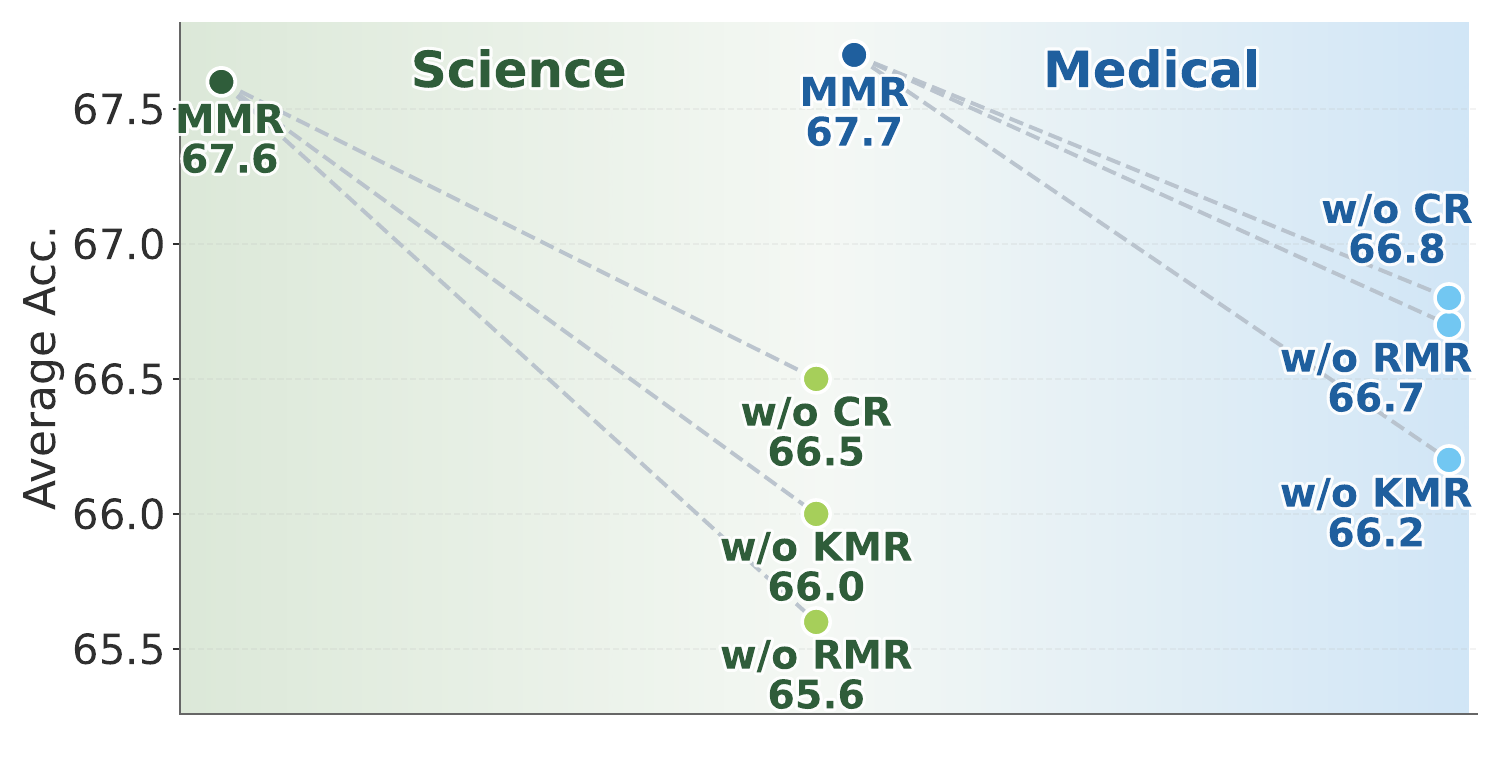}
\caption{
\textbf{Effect of different reward components.}
We ablate KMR, RMR, and CR from the reward function and evaluate the resulting models.
The figure reports average accuracy under each ablation setting compared with the full reward.
}
\label{fig:ablation_reward}
\end{minipage}
\end{figure*}
\noindent\textbf{Effect of grader choice.}
Figure~\ref{fig:ablation_grader} analyzes the effect of grader choice. 
We compare three graders, including \mqwen, \oss, and \xxlqwen, and report the average accuracy on science and medical benchmarks. 
The dashed lines denote the corresponding \mqwen~baseline without \sys. 
We find that:

\noindent\textbf{Stronger graders provide more effective metacognitive supervision.}
Among the three graders, \xxlqwen~achieves the best performance in both domains, reaching 67.6\% on science and 67.7\% on medical.
Compared with the \mqwen~baseline, it improves the average accuracy by 3.8\% and 1.7\%, respectively.
This suggests that stronger graders provide higher-quality process supervision, leading to more effective policy optimization.

\noindent\textbf{\sys remains effective even with a lightweight self-grader.}
Using \mqwen~as the grader still improves over the \mqwen~baseline, increasing science performance from 63.8\% to 65.9\% and medical performance from 66.0\% to 66.6\%.
This indicates that \sys~does not strictly rely on a very large external grader.
Even when the policy is supervised by a grader of the same scale, metacognitive supervision still brings consistent gains, making \sys~more practical for resource-constrained settings.

\noindent \textbf{Effect of reward component.}
Figure~\ref{fig:ablation_reward} studies the effect of each reward component by removing KMR, RMR, and CR from \sys respectively. 
We report the average accuracy on science and medical benchmarks. We observe that:

\noindent\textbf{All reward components contribute to \sys, with process-level supervision playing a particularly important role.}
Removing any component leads to performance degradation on both science and medical benchmarks, showing that KMR, RMR, and CR are all useful for effective optimization.
On science benchmarks, removing CR, KMR, and RMR decreases the average accuracy from 67.6\% to 66.5\%, 66.0\%, and 65.6\%, respectively.
On medical benchmarks, the corresponding scores drop from 67.7\% to 66.8\%, 66.2\%, and 66.7\%.
Notably, removing KMR or RMR generally causes larger degradation than removing CR, suggesting that process-level supervision through knowledge monitoring and regulation fidelity plays a central role in the effectiveness of \sys.

\section{Conclusion and Limitation}

\noindent \textbf{Conclusion.}
In this work, we introduce \sys, a metacognition-inspired RL framework that uses metacognitive knowledge and regulation as general reward dimensions for LLM reasoning. \sys structures rollouts into explicit metacognitive components and optimizes them with a trajectory-level reward over knowledge coverage, regulation fidelity, and answer correctness. By grounding reward design in general metacognitive dimensions, \sys extends reward feedback from final-answer outcomes to reasoning trajectories, while avoiding the need for hand-crafted instance-specific rubrics. Experiments on 22 benchmarks show consistent gains in task performance, reasoning process quality, and out-of-domain generalization. 

\noindent \textbf{Limitation.}
Despite our best efforts to formulate \sys as a general metacognitive reward framework and evaluate it across a broad set of reasoning benchmarks, our work still has several limitations. First, although our evaluation covers multiple reasoning domains, the training data are mainly derived from science and medical tasks, and broader training domains may further improve generality. Second, our experiments focus on text-only reasoning, leaving multilingual, multimodal, and interactive agent settings for future exploration.

\bibliography{iclr2026_conference}
\bibliographystyle{iclr2026_conference}

\clearpage
\appendix

\section{Prompt Template}

We provide the prompt templates for rollout generation, grader scoring, and data generation in Figures~\ref{fig:appendix_rollout}, \ref{fig:appendix_grader}, and \ref{fig:data_generation}, respectively.

\begin{tcolorbox}[
    enhanced,
    breakable,
    width=\linewidth,
    colback=softgray,
    colframe=teal!30,
    coltitle=black,
    title=\textbf{Rollout Generation Prompt Template},
    boxrule=0.9pt,
    arc=1.5mm,
    left=2mm,
    right=2mm,
    top=1mm,
    bottom=1mm,
    before skip=8pt,
    after skip=8pt
]
\begin{Verbatim}[
    fontsize=\footnotesize,
    breaklines=true,
    breakanywhere=true,
    breaksymbolleft={},
    breaksymbolright={}
]
In a woman with mitral stenosis but no pulmonary hypertension who is in labor at 39 weeks of gestation, which medical intervention should be avoided to prevent complications associated with her heart condition during the third stage of labor?

========= SOLVE STRATEGY =========
Above you are given a task. Follow the two steps to solve the task.

Step 1 -- Meta thinking
Identify any knowledge that is task-relevant, necessary, or highly important for solving the task.

Metacognitive Knowledge (MK): task-relevant knowledge needed for solving
(e.g., key facts / definitions / constraints / rules / procedures / exceptions).
- [mk1] 
- [mk2]

Metacognitive Regulation (MR): 
- [plan] # short executable steps

Then solve using the MK and MR.
If you become blocked, uncertain, or realize that an important piece of knowledge was not included in MK, trigger LOOKBACK.

LOOKBACK:
- [Seeking] # exactly what you need
- [Found] # the recalled or derived knowledge
Then continue solving.

Step 2 -- Answer
Check: does your answer apply the necessary knowledge and follow the correct plan?
Provide the final answer.

Rules:
- The meta thinking process and answer must be enclosed within <meta> and <answer> tags, respectively. 

Output format:
<meta>meta thinking process</meta><answer>final answer</answer>
\end{Verbatim}
\end{tcolorbox}

\captionof{figure}{
\textbf{Prompt template for rollout generation.}
The prompt asks the policy $\pi_\theta$ to generate the MK, MR, optional lookback and final answer.
}
\label{fig:appendix_rollout}

\begin{tcolorbox}[
    enhanced,
    breakable,
    width=\linewidth,
    colback=softgray,
    colframe=teal!30,
    coltitle=black,
    title=\textbf{Grader Scoring Prompt Template},
    boxrule=0.9pt,
    arc=1.5mm,
    left=2mm,
    right=2mm,
    top=1mm,
    bottom=1mm,
    before skip=8pt,
    after skip=8pt
]
\begin{Verbatim}[
    fontsize=\footnotesize,
    breaklines=true,
    breakanywhere=true,
    breaksymbolleft={},
    breaksymbolright={}
]
You are a strict evaluator for scoring.

Your task is to evaluate the actor's output and return exactly five values:
- k: the number of gold knowledge units covered by the actor's metacognitive knowledge (MK)
- r: the number of initially missing gold knowledge units recovered through LOOKBACK
- a: an alignment score in [0,1] measuring how well the actor's final answer follows the actor's own metacognitive regulation (MR)
- s: a shortcut flag in {{0,1}}
- c: a correctness flag in {{0,1}}

================================================
SECTION IDENTIFICATION
================================================

[Model Output] may contain MK, MR, and LOOKBACK. Matching is case-insensitive.
- MK: any section labeled as MK, mk, or "Metacognitive Knowledge"
- MR: any section labeled as MR, mr, or "Metacognitive Regulation"
- LOOKBACK: any section labeled as LOOKBACK or lookback

If a section is absent or cannot be clearly identified by its label, treat it as missing and apply the missing-section rules defined below.

================================================
DEFINITIONS AND SCORING RULES
================================================

------------------------------------------------
1. Gold Knowledge
------------------------------------------------
These are the only references for evaluating k and r. Each numbered item is exactly one gold knowledge unit.

------------------------------------------------
2. k -- Covered Gold Knowledge
------------------------------------------------
A gold knowledge unit is covered if the actor's MK expresses the same semantic content, even if worded differently.
- Do NOT count vague or partial matches.
- Do NOT invent new gold knowledge units.
- Coverage is binary: a unit is either covered or not. No partial credit.

------------------------------------------------
3. r -- Recovered Missing Gold Knowledge
------------------------------------------------
Step 1: Identify which gold knowledge units are absent from MK.
Step 2: Inspect the LOOKBACK section.
Count a unit as recovered ONLY if:
  - it was absent from MK, AND
  - the same semantic content appears in LOOKBACK.
- Recovery is binary: a unit is either recovered or not. No partial credit.
Note: 0 <= r <= (Number of Gold Knowledge - k)

------------------------------------------------
4. a -- Regulation-Answer Alignment
------------------------------------------------
Score ONLY the consistency between [Model Final Answer] and MR.
Do NOT score whether the final answer is factually correct.

Use the following as anchor points. Prefer anchor values when the case clearly fits one description. You may use intermediate values (e.g., 0.6, 0.8) only when the case falls between two anchors and neither fits well.

  1.0  -- [Model Final Answer] clearly and fully executes MR
  0.75 -- [Model Final Answer] mostly executes MR, with only minor deviation
  0.5  -- [Model Final Answer] partially executes MR
  0.25 -- [Model Final Answer] weakly reflects MR
  0.0  -- [Model Final Answer] does not meaningfully follow MR

------------------------------------------------
5. s -- Shortcut Flag
------------------------------------------------
Set s = 1 only if there is clear evidence that the actor bypasses its own visible metacognitive process and jumps directly to the final answer. Do not set s = 1 merely because the response is brief or lacks one labeled section.

------------------------------------------------
6. c -- Final Answer Correctness
------------------------------------------------
Compare ONLY [Model Final Answer] with [Ground Truth].
Set c = 1 if the actor's final answer is identical to or semantically equivalent to the Ground Truth. Set c = 0 otherwise.

------------------------------------------------
7. Additional rules
------------------------------------------------
- If there is no identifiable MK section, set k = 0.
- If there is no valid LOOKBACK section, set r = 0.
- If there is no identifiable MR or no identifiable final answer, score a conservatively.
- Ground Truth must NOT be used to directly score k, r, a, or s. It may be used only for scoring c.

================================================
INPUT
================================================

[Gold Knowledge Units]
{gold_knowledge_text}

[Number of Gold Knowledge]
{number_of_gold_knowledge}

[Ground Truth]
{ground_truth_text}

[Model Output]
{model_output}

[Model Final Answer]
{model_final_answer}

================================================
OUTPUT
================================================

Return JSON only. Do not include markdown, prose, or any text outside the JSON object.

{{
  "k": <integer, 0 to {number_of_gold_knowledge}>,
  "r": <integer, 0 to ({number_of_gold_knowledge} - k)>,
  "a": <float, 0.0 to 1.0>,
  "s": <0 or 1>,
  "c": <0 or 1>
}}
\end{Verbatim}
\end{tcolorbox}

\captionof{figure}{
\textbf{Prompt template for grader scoring.}
The prompt instructs the grader to assess whether each rollout from $\pi_\theta$ covers the required metacognitive knowledge, follows its stated regulation plan, and produces a correct final answer.
}\label{fig:appendix_grader}

\begin{tcolorbox}[
    enhanced,
    breakable,
    width=\linewidth,
    colback=softgray,
    colframe=teal!30,
    coltitle=black,
    title=\textbf{Data Generation Prompt Template},
    boxrule=0.9pt,
    arc=1.5mm,
    left=2mm,
    right=2mm,
    top=1mm,
    bottom=1mm,
    before skip=8pt,
    after skip=8pt
]
\begin{Verbatim}[
    fontsize=\footnotesize,
    breaklines=true,
    breakanywhere=true,
    breaksymbolleft={},
    breaksymbolright={},
    commandchars=\\\|\!
]
\textbf|System Prompt:!

You are constructing supervision metadata for metacognitive RL.

Your task is to extract supervision metadata from the given task and reference answer.

Normal case:
1. gold_knowledge
A set of atomic gold metacognitive knowledge (e.g., key facts / definitions / constraints / rules / procedures / exceptions) required for solving the task.

2. possible_meta_regulation
One regulation description that captures a reasonable solving process.

Important constraints:
- Use the task as the primary source.
- Use the reference answer only as auxiliary reference for relevance and necessity.
- Do NOT simply paraphrase or decompose the reference answer into trivial answer-support bullets.
- gold_knowledge units must be:
  - atomic and self-contained
  - task-relevant, necessary or highly important for solving
  - collectively cover all necessary information required to solve the task
  - non-redundant and minimally overlapping
  - specific enough to be useful, and not overly generic
- possible_meta_regulation should be:
  - exactly one entry with id "PMR1"
  - high-level and process-oriented
  - not overly tied to the exact wording of the reference answer

Special rule:
- If you judge that this task and its reference answer are not suitable for constructing trustworthy supervision metadata, then do NOT generate gold_knowledge.
- This includes cases such as: the task or reference answer being unreliable, malformed, self-contradictory, missing key information, obviously wrong, containing minor errors, being logically not rigorous, ambiguous in a way that affects solvability, or otherwise unsuitable for high-quality supervision construction.
- In that case, return ONLY:
  {
    "possible_meta_regulation": [{"id": "PMR1", "text": "error"}]
  }

Return JSON only.

\textbf|User Prompt:!

Extract the required fields for the following sample.

Return JSON only in one of the following two formats.

Normal case:
{
  "gold_knowledge": [
    {"id": "KU1", "text": "..."},
    {"id": "KU2", "text": "..."}
  ],
  "possible_meta_regulation": [
    {"id": "PMR1", "text": "..."}
  ]
}

If you believe this task and its reference answer are not suitable for trustworthy supervision construction, return ONLY:
{
  "possible_meta_regulation": [
    {"id": "PMR1", "text": "error"}
  ]
}

[Task]
{task_prompt}

[Reference Answer]
{ground_truth}

\end{Verbatim}
\end{tcolorbox}

\captionof{figure}{
\textbf{Prompt template for data generation.}
The prompt asks the model to generate both gold knowledge and possible meta regulation. The latter is an auxiliary field and is not used during training.
}
\label{fig:data_generation}

\section{Dataset Details}
We present the details of the training set in Figures~\ref{fig:data_medical} and~\ref{fig:data_science}.

\begin{tcolorbox}[
    enhanced,
    breakable,
    width=\linewidth,
    colback=softgray,
    colframe=teal!30,
    coltitle=black,
    title=\textbf{Training Data Example (Medical)},
    boxrule=0.9pt,
    arc=1.5mm,
    left=2mm,
    right=2mm,
    top=1mm,
    bottom=1mm,
    before skip=8pt,
    after skip=8pt
]
\begin{Verbatim}[
    fontsize=\footnotesize,
    breaklines=true,
    breakanywhere=true,
    breaksymbolleft={},
    breaksymbolright={},
]
{
  "sample_id": "sample_11180",
  "prompt": [
    {
      "role": "user",
      "content": "A one-year-old child presents with short stature, lethargy, and constipation, along with a palpable goiter. Laboratory tests show low T4 and elevated TSH levels. What is the most likely diagnosis for this child's condition?\n\n========= SOLVE STRATEGY =========\nAbove you are given a task. Follow the two steps to solve the task.\n\n## Step 1 — Meta thinking \nIdentify any knowledge that is task-relevant, necessary, or highly important for solving the task.\n\nMetacognitive Knowledge (MK): task-relevant knowledge needed for solving\n(e.g., key facts / definitions / constraints / rules / procedures / exceptions).\n- [mk1] \n- [mk2]\n\nMetacognitive Regulation (MR): \n- [plan] # short executable steps\n\nThen solve using the MK and MR.\nIf you become blocked, uncertain, or realize that an important piece of knowledge was not included in MK, trigger LOOKBACK.\n\nLOOKBACK:\n- [Seeking] # exactly what you need\n- [Found] # the recalled or derived knowledge\nThen continue solving.\n\n## Step 2 — Answer \nCheck: does your answer apply the necessary knowledge and follow the correct plan?\nProvide the final answer.\n\nRules:\n- The meta thinking process and answer must be enclosed within <meta> and <answer> tags, respectively. \n \nOutput format:\n<meta>meta thinking process</meta><answer>final answer</answer>."
    }
  ],
  "data_source": "FreedomIntelligence/medical-o1-reasoning-SFT",
  "ability": "medical-qa",
  "reward_model": {
    "style": "metacognitive",
    "ground_truth": "The most likely diagnosis for this child's condition is congenital hypothyroidism due to dyshormonogenesis. The symptoms of short stature, lethargy, constipation, and a palpable goiter, combined with laboratory findings of low T4 and elevated TSH, suggest that the thyroid gland is present but not functioning properly due to an issue in hormone synthesis. This aligns with dyshormonogenesis, where there is a defect in the process of producing thyroid hormones, leading to primary hypothyroidism and compensatory goiter development.",
    "gold_knowledge_num": 5,
    "gold_knowledge": [
      {
        "id": "KU1",
        "text": "In primary hypothyroidism, serum T4 levels are low and TSH levels are elevated due to loss of negative feedback."
      },
      {
        "id": "KU2",
        "text": "A palpable goiter in a hypothyroid child indicates that thyroid tissue is present and enlarged rather than absent or ectopic."
      },
      {
        "id": "KU3",
        "text": "Dyshormonogenesis is a congenital defect in thyroid hormone synthesis that typically presents with a goitrous form of congenital hypothyroidism."
      },
      {
        "id": "KU4",
        "text": "Clinical features of hypothyroidism in infants and young children include lethargy, constipation, and growth failure or short stature."
      },
      {
        "id": "KU5",
        "text": "Congenital hypothyroidism caused by thyroid dysgenesis (agenesis or ectopy) usually presents without a palpable goiter because little or no thyroid tissue is present."
      }
    ],
    "possible_meta_regulation": [
      {
        "id": "PMR1",
        "text": "Identify the syndrome suggested by the clinical features and labs (hypothyroidism with low T4 and high TSH), then determine whether the thyroid gland is present or absent based on the presence of a goiter, and use this to distinguish among causes of congenital hypothyroidism, selecting dyshormonogenesis when there is a goitrous, primary hypothyroidism due to defective hormone synthesis."
      }
    ]
  },
  "extra_info": ...
}

\end{Verbatim}
\end{tcolorbox}

\captionof{figure}{
\textbf{Training data example from the medical domain.}
The example illustrates the input prompt and the corresponding metacognitive supervision signals used for training.
}
\label{fig:data_medical}
\begin{tcolorbox}[
    enhanced,
    breakable,
    width=\linewidth,
    colback=softgray,
    colframe=teal!30,
    coltitle=black,
    title=\textbf{Training Data Example (Science)},
    boxrule=0.9pt,
    arc=1.5mm,
    left=2mm,
    right=2mm,
    top=1mm,
    bottom=1mm,
    before skip=8pt,
    after skip=8pt
]
\begin{Verbatim}[
    fontsize=\footnotesize,
    breaklines=true,
    breakanywhere=true,
    breaksymbolleft={},
    breaksymbolright={},
]
{
  "sample_id": "sample_8131",
  "prompt": [
    {
      "role": "user",
      "content": "How can genetic mutations produce new organs or features, and what are the conditions under which these mutations can be beneficial or harmful to an organism? Provide examples of vestigial organs and explain how they relate to the concept of evolution.\n\n========= SOLVE STRATEGY =========\nAbove you are given a task. Follow the two steps to solve the task.\n\n## Step 1 — Meta thinking \nIdentify any knowledge that is task-relevant, necessary, or highly important for solving the task.\n\nMetacognitive Knowledge (MK): task-relevant knowledge needed for solving\n(e.g., key facts / definitions / constraints / rules / procedures / exceptions).\n- [mk1] \n- [mk2]\n\nMetacognitive Regulation (MR): \n- [plan] # short executable steps\n\nThen solve using the MK and MR.\nIf you become blocked, uncertain, or realize that an important piece of knowledge was not included in MK, trigger LOOKBACK.\n\nLOOKBACK:\n- [Seeking] # exactly what you need\n- [Found] # the recalled or derived knowledge\nThen continue solving.\n\n## Step 2 — Answer \nCheck: does your answer apply the necessary knowledge and follow the correct plan?\nProvide the final answer.\n\nRules:\n- The meta thinking process and answer must be enclosed within <meta> and <answer> tags, respectively. \n \nOutput format:\n<meta>meta thinking process</meta><answer>final answer</answer>."
    }
  ],
  "data_source": "Meta/natural_reasoning",
  "ability": "science-qa",
  "reward_model": {
    "style": "metacognitive_rubric",
    "ground_truth": "Genetic mutations can produce new organs or features through various mechanisms, such as saltations, which are mutations with a large effect on morphology. These mutations can be beneficial or harmful depending on the environment, and examples of vestigial organs, such as webbed feet, can illustrate this concept.",
    "gold_knowledge_num": 9,
    "gold_knowledge": [
      {
        "id": "KU1",
        "text": "Genetic mutations are changes in an organism’s DNA sequence that can alter development and morphology, potentially producing new or modified organs or features over evolutionary time."
      },
      {
        "id": "KU2",
        "text": "New organs or complex features typically arise not from a single mutation but from the cumulative effect of many mutations that modify existing structures or developmental pathways."
      },
      {
        "id": "KU3",
        "text": "Large-effect mutations (sometimes called saltational changes) can cause major morphological shifts, but most mutations are small in effect and contribute incrementally to phenotypic change."
      },
      {
        "id": "KU4",
        "text": "Whether a mutation is beneficial, neutral, or harmful depends on the specific environment: a change that increases reproductive success in a given environment is beneficial, one that decreases it is harmful, and one with no effect on fitness is neutral."
      },
      {
        "id": "KU5",
        "text": "Natural selection tends to increase the frequency of beneficial mutations and decrease the frequency of harmful ones, whereas neutral mutations can drift in frequency due to genetic drift."
      },
      {
        "id": "KU6",
        "text": "Vestigial organs are structures that have lost most or all of their original function in a species but are homologous to fully functional structures in related species."
      },
      {
        "id": "KU7",
        "text": "Examples of vestigial structures in humans include the appendix, the coccyx (tailbone), wisdom teeth, and the muscles that move the ears; in other animals, examples include the pelvic bones in whales and snakes and reduced wings in flightless birds."
      },
      {
        "id": "KU8",
        "text": "Vestigial organs provide evidence for evolution because they indicate descent from ancestors in which these structures were functional, supporting the idea of modification of pre-existing structures over time."
      },
      {
        "id": "KU9",
        "text": "The same mutation can be beneficial in one environmental or ecological context and harmful in another, illustrating the context-dependence of selection on new traits."
      }
    ],
    "possible_meta_regulation": [
      {
        "id": "PMR1",
        "text": "Break the task into subtasks: (1) explain mechanistically how mutations can change morphology and give rise to new or modified organs, noting both cumulative small changes and possible large-effect mutations; (2) analyze the conditions under which such mutations count as beneficial, harmful, or neutral using fitness and environment as criteria, and illustrate with at least one contextual example; (3) define vestigial organs precisely, list several concrete examples from humans and other animals, and (4) explicitly connect vestigial structures to evolutionary theory by explaining how they serve as evidence for descent with modification. Ensure each part of the question is answered and that the examples logically support the conceptual explanations."
      }
    ]
  },
  "extra_info": ...
}

\end{Verbatim}
\end{tcolorbox}

\captionof{figure}{
\textbf{Training data example from the science domain.}
The example illustrates the input prompt and the corresponding metacognitive supervision signals used for training.
}
\label{fig:data_science}

\section{Detailed Training Settings}
\label{appendix_training_details}
For RL training, we set both the maximum prompt length and response length to 4096 tokens, with an overlong buffer of 256 tokens enabled, no KL reward/loss.
The lower and upper clip ratios are to 0.2 and 0.28, respectively.
During rollout generation, we use temperature $=1.0$, top-$p=1.0$, and top-$k=-1$.
The actor learning rate is $2\times 10^{-6}$ with 10 warmup steps.
We use a training batch size of 224, generate $N=8$ rollouts per prompt, set the mini-batch size to 112, and train for 270 steps. All experiments are conducted with NVIDIA H200 GPUs, each with 141GB of memory.

\section{Qualitative Analysis}

Figure~\ref{fig:appendix_policy} shows a structured rollout generated by $\pi_\theta$ during training.
\begin{tcolorbox}[
    enhanced,
    breakable,
    width=\linewidth,
    colback=softgray,
    colframe=teal!30,
    coltitle=black,
    title=\textbf{Example of Generated Rollout},
    boxrule=0.9pt,
    arc=1.5mm,
    left=2mm,
    right=2mm,
    top=1mm,
    bottom=1mm,
    before skip=8pt,
    after skip=8pt
]
\begin{Verbatim}[
    fontsize=\footnotesize,
    breaklines=true,
    breakanywhere=true,
    breaksymbolleft={},
    breaksymbolright={},
    commandchars=\\\{\}
]
\textbf{Input:}

Diethyl ether reacts with sodium metal to give (a) Ethanol (b) Sodium ethoxide (c) Ethanal (d) Nothing happens

--Omitted the SOLVE STRATEGY Part--

\textbf{Policy Generation:}

<meta>
## Step 1 \textemdash{} Meta Thinking

### Metacognitive Knowledge (MK)

**MK1 - Key Facts about Diethyl Ether and Sodium Metal Reaction:**

* Diethyl ether has the structure CH\textsubscript{3}CH\textsubscript{2}-O-CH\textsubscript{2}CH\textsubscript{3} (C\textsubscript{2}H\textsubscript{5}OC\textsubscript{2}H\textsubscript{5}) - an ether functional group with no O-H bond
* Sodium metal (Na) reacts vigorously with compounds containing acidic/hydrogen atoms attached directly to oxygen (like alcohols: R-OH)
* Alcohols react with Na: 2 R-OH + 2 Na \(\rightarrow\) 2 R-ONa (alkoxide) + H\textsubscript{2}(g)
* Ethers lack acidic protons - hydrogens are attached to carbon, not oxygen, making them very weakly acidic
* Simple alkyl ethers are chemically inert to sodium metal under normal conditions
* Ethers react with Sodium primarily through H abstraction from trace moisture or impurities if any, not from the ether molecule itself

**MK2 - Comparison with Related Compounds:**

* Ethanol (C\textsubscript{2}H\textsubscript{5}OH) + Na \(\rightarrow\) Sodium ethoxide (C\textsubscript{2}H\textsubscript{5}ONa) + H\textsubscript{2} (this works)
* Diethyl ether (C\textsubscript{2}H\textsubscript{5}OC\textsubscript{2}H\textsubscript{5}) + Na \(\rightarrow\) No reaction (no acidic H to give up)
* Ethanal is an aldehyde, not produced from this reaction
* The question tests understanding that ethers are much less reactive than alcohols due to lack of O-H bond
* In laboratory practice, diethyl ether is often used as a solvent precisely because it is unreactive with alkali metals like Na, K, Li

**MK3 - Reaction Constraints and Special Cases:**

* Some very reactive conditions at high temperature or with specific metal amide bases could cleave ethers, but not with Na metal alone
* Autoxidation of ethers with -O-O-H impurities can occur with air + Na, but not the ether itself reacting directly
* Sodium reacts with ether impurities (water/ethanol) but if pure diethyl ether + Na, nothing happens
* The classic organic chemistry rule: "Alcohols react with Na, Ethers don't - this is a key distinction"

**MK4 - Exam/Question Pattern Recognition:**

* This is a standard objective question testing functional group reactivity differences
* The correct answer is commonly taught in organic chemistry courses around ether stability
* Options (a), (b), (c) represent reactions that require specific conditions or different reactants
* Option (d) "Nothing happens" reflects the known inertness of ethers toward Na metal

### Metacognitive Regulation (MR)

* **Plan:**

  1. Analyze the functional groups present (ether vs alcohol)
  2. Recall sodium's reactivity profile (reacts with protic compounds, not hydrocarbons/ethers)
  3. Identify that diethyl ether has no acidic hydrogen for Na to deprotonate
  4. Eliminate all options that require different reactants/mechanisms
  5. Conclude "Nothing happens" is correct based on ether inertness
* **If blocked:** Check if special conditions exist (answer confirms standard conditions apply, no special cases mentioned)
* **Noticing:** The answer seems counterintuitive to students expecting a reaction - this is the teaching point of the question

### LOOKBACK:

* **Seeking:** Is there any exception where sodium cleaves diethyl ether?
* **Found:** Sodium can eventually react with ethers only under extreme conditions or via radical mechanisms not relevant here. Standard undergraduate chemistry teaches ethers are inert to Na metal. Lithium metal in certain conditions can cleave ethers (Williamson-type), but Na alone does not. This confirms option (d) is correct.

</meta>

<answer>
The correct answer is **(d) Nothing happens**.

**Explanation:**
Diethyl ether (C\textsubscript{2}H\textsubscript{5}OC\textsubscript{2}H\textsubscript{5}) is an ether with no O-H (hydroxyl) protons. Sodium metal reacts with compounds containing acidic hydrogens attached to oxygen (such as alcohols) to form alkoxides and hydrogen gas:

2 C\textsubscript{2}H\textsubscript{5}OH + 2 Na \(\rightarrow\) 2 C\textsubscript{2}H\textsubscript{5}ONa (sodium ethoxide) + H\textsubscript{2}\(\uparrow\)

However, **diethyl ether lacks this acidic hydrogen**. The protons in ether are attached to carbon atoms, not oxygen, making them extremely weakly acidic. Ethers are known for their chemical inertness and are commonly used as solvents that do not react with alkali metals like sodium under normal conditions.

* (a) Ethanol would not form from ether + Na
* (b) Sodium ethoxide would form from ethanol + Na, not ether + Na
* (c) Ethanal is an aldehyde, not produced in this reaction
* (d) Nothing happens - \(\checkmark\) correct because ethers are inert to sodium metal

**Answer: (d)** 
</answer>

\end{Verbatim}
\end{tcolorbox}

\captionof{figure}{
\textbf{Example generation from the policy $\pi_\theta$.}
During training, $\pi_\theta$ can organize its response into the required metacognitive format.
For the given question, it extracts relevant metacognitive knowledge, provides a detailed reasoning plan in metacognitive regulation, triggers \texttt{LOOKBACK} under uncertainty, and produces a faithful and correct final answer.
}
\label{fig:appendix_policy}

\section{Broader Impacts}
This work may have positive societal impacts by improving the reliability of LLM reasoning through process-oriented reward design. By encouraging models to identify task-relevant knowledge and follow explicit reasoning plans, \sys may support more interpretable and controllable reasoning behavior. At the same time, stronger reasoning models and LLM-based reward systems may still produce incorrect outputs. Therefore, \sys should be viewed as a research framework rather than a deployment-ready safety mechanism, and practical use should involve careful validation.

\end{document}